\documentclass{article}

\usepackage{arxiv}
\usepackage[utf8]{inputenc} 
\usepackage[T1]{fontenc}    
\usepackage[hidelinks]{hyperref}      
\usepackage{cleveref}
\usepackage{url}            
\usepackage{booktabs}       
\usepackage{amsfonts}       
\usepackage{nicefrac}       
\usepackage{microtype}      
\usepackage{lipsum}
\usepackage{xcolor}
\usepackage{amssymb} 
\usepackage{graphicx}
\graphicspath{ {./figures/} } 
\usepackage{array}
\newcolumntype{L}[1]{>{\raggedright\let\newline\\\arraybackslash\hspace{0pt}}m{#1}}
\newcolumntype{C}[1]{>{\centering\let\newline\\\arraybackslash\hspace{0pt}}m{#1}}
\newcolumntype{R}[1]{>{\raggedleft\let\newline\\\arraybackslash\hspace{0pt}}m{#1}}
\usepackage{setspace}
\usepackage{authblk}
\author[1]{\large Hian Hian See}
\author[2]{\large Brian Lim}
\author[1,2]{\large Si Li}
\author[1]{\large Haicheng Yao}
\author[1]{\large Wen Cheng}
\author[4]{\large Harold Soh}
\author[1,2,3]{\large Benjamin C.K. Tee}

\affil[1]{\footnotesize	 Department of Materials Science and Engineering, National University of Singapore}
\affil[2]{\footnotesize	 Institute for Health Innovation and Technology, National University of Singapore}
\affil[3]{\footnotesize Department of Electrical Engineering, National University of Singapore}
\affil[4]{\footnotesize Department of Computer Science, National University of Singapore}
\affil[ ]{\texttt{Email: harold@comp.nus.edu.sg, benjamin.tee@nus.edu.sg}}

\title{ST-MNIST - The Spiking Tactile-MNIST Neuromorphic Dataset}
\begin{document}
\maketitle
\begin{abstract}

Tactile sensing is an essential modality for smart robots as it enables them to interact flexibly with physical objects in their environment. Recent advancements in electronic skins have led to the development of data-driven machine learning methods that exploit this important sensory modality. However, current datasets used to train such algorithms are limited to standard synchronous tactile sensors. There is a dearth of \emph{neuromorphic} event-based tactile datasets, principally  due to the scarcity of large-scale event-based tactile sensors. Having such datasets is crucial for the development and evaluation of new algorithms that process spatio-temporal event-based data. For example, evaluating spiking neural networks on conventional frame-based datasets is considered sub-optimal. Here, we debut a novel neuromorphic Spiking Tactile MNIST (ST-MNIST) dataset, which comprises   handwritten digits obtained by human participants writing on a neuromorphic tactile sensor array. We also describe an initial effort to evaluate our ST-MNIST dataset using existing artificial and spiking neural network models. The classification accuracies provided herein can serve as performance benchmarks for future work. We anticipate that our ST-MNIST dataset will be of interest and useful to the neuromorphic and robotics research communities.

\end{abstract}

\newpage
\section{Introduction} \label{introduction}

Humans use touch as an intuitive sense to perceive and understand our physical world. In robotics, tactile sensing has received increasing attention since the early 1980s~\cite{nicholls1989survey,Siciliano2017,Hammock2013}. Tactile sensing via an electronic skin augments the robot’s perception of the physical world with information beyond what standard vision and auditory modalities can provide. For instance, touch sensing allows robots to perceive objects based on their physical properties, e.g., surface texture, weight, and stiffness. Touch is also essential when vision is obscured during manipulation. The advent of an artificial skin that rivals the human analog opens up tremendous potential in robotics, and opportunities extend to other domains such as prosthetics~\cite{Li2020} and human-machine interaction~\cite{cao2018screen,lopes2018hydroprinted}.
Coupled with the rapid advancement of machine learning, new tactile sensors will play a major role in autonomous robots capable of deft manipulation in  (unstructured) physical spaces. 

Artificial skins provide a vast amount of data and data-driven machine learning has emerged as a key approach for modeling the complexities associated with touch. For example, recent work~\cite{sundaram2019learning} recorded a large dataset of 135,000 tactile maps using a tactile glove (with 548 pressure sensors). Convolution neural networks (CNNs) were used to learn the tactile signatures and accurately identify the class and weight of the grasped objects.  However, many works~\cite{sundaram2019learning,taunyazov2019towards,larson2019deformable,alameh2020smart,soh2012online,Soh2014} on smart tactile sensing are based on standard synchronous tactile sensors, where individual taxels are sampled sequentially and periodically to construct pressure distribution maps. The serial readout nature of these synchronous sensors results in readout latency bottlenecks as the number of sensors increases. This drawback slows down the transfer of tactile stimuli to the learning algorithms for inference, which could be a major obstacle for real-world applications where machines are required to respond swiftly.  

A more efficient way of capturing large amounts of tactile data is via \emph{event-based} tactile sensors~\cite{lee2019neuro}. Unlike synchronous sensors, event-based sensors are not periodically polled by a central electronic controller. Instead, tactile information is transmitted asynchronously as \emph{events} --- digital packets of data that carry information about a change in pressure, its location, and a timestamp. Since only changes are captured, event-sensors promise higher energy-efficiency, lower readout latency, and higher temporal resolution. However,  event-based tactile sensors remain under-developed compared to their synchronous counterparts. The scarcity of the event-based tactile sensors has caused a dearth of event-based tactile datasets, which hampers the development and evaluation of new machine learning algorithms that can exploit the rich spatio-temporal features of touch data~\cite{Pfeiffer2018,liu2019event}.

To fill this gap, we debut a novel neuromorphic Spiking Tactile MNIST (ST-MNIST) dataset.  ST-MNIST comprises handwritten digits (0-9) generated by 23 human subjects writing on a 100-taxel biomimetic event-based tactile sensor array~\cite{lee2019neuro}. The closest related dataset is the neuromorphic vision-based N-MNIST~\cite{orchard2015converting}, where event-data was ``artificially'' obtained via conversion from static images using an event camera (i.e., ATIS~\cite{posch2010qvga}). In contrast, ST-MNIST captures the motion and pressure dynamics associated with natural writing. We have made ST-MNIST publicly available\footnote{\url{www.benjamintee.com/stmnist}} to encourage research on tactile perception. In this paper, we describe performance evaluations of ST-MNIST using both Artificial (ANN) and Spiking Neural Network (SNN) models to serve as performance benchmarks. 

The rest of the paper is organized as follows: we start off by describing our data collection process in Section \ref{methods}. In Section \ref{data_properties}, we show some basic properties of the ST-MNIST dataset. In Section \ref{classification}, we present the classification results of the ST-MNIST dataset using existing ANN and SNN models, before concluding in Section \ref{conclusion}. 
 
\section{Materials and Methods} \label{methods}
In this section, we present our data collection methodology. We first describe the design of our neuromorphic tactile sensing system (Section \ref{tactilesystem}), followed by the data collection process (Section \ref{datacollection}). We also provide details on how to acquire and interpret the resulting dataset (Section \ref{fileformat}).

\subsection{Design of Neuromorphic Tactile System} \label{tactilesystem}
Our neuromorphic tactile system utilizes an asynchronous signaling system with an array of tactile transducers. The event-driven system --- Asynchronously Coded Electronic Skin (ACES) --- is akin to the human peripheral nervous system~\cite{lee2019neuro}. It was developed to address the increasing complexity and need for transferring large array of skin-like transducer inputs while maintaining a high level of responsiveness (i.e., low latency). 

\begin{figure}
	\centering
	\includegraphics[scale=0.3]{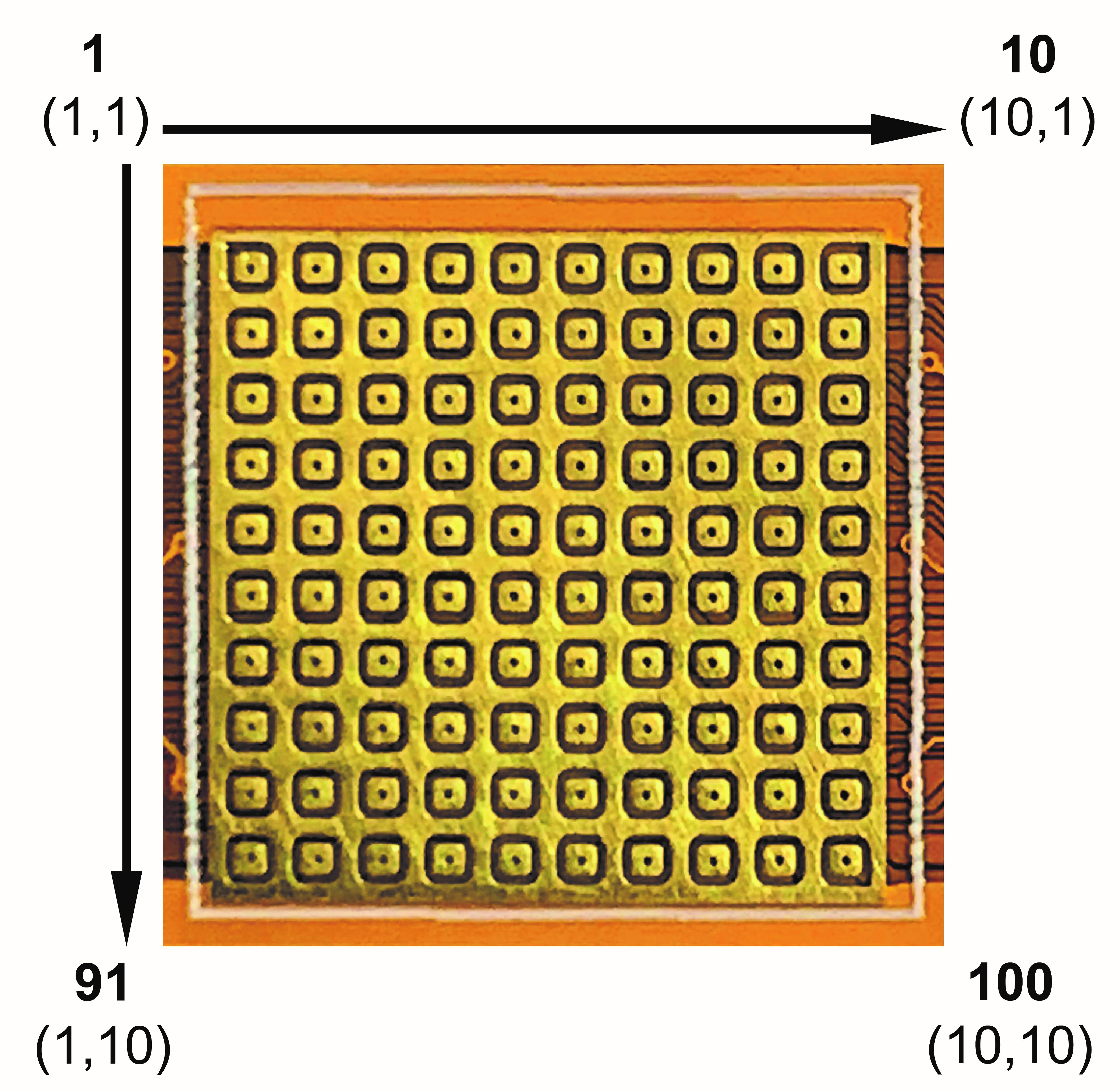}
	\caption{Spatial distribution of the 100 taxels on the tactile sensor array. The taxel addresses are in bold, and the X and Y addresses are provided in brackets, i.e., (X,Y).}
	\label{fig:electrode100}
\end{figure}

In this work, our tactile sensing is realized via a high-density electrode array of 100 taxels (tactile pixels) cm$^{-2}$ (Fig. \ref{fig:electrode100}) and a piezoresistive thin film (i.e., pressure transducer). We interfaced the tactile sensor array with the ACES platform for the creation of the neuromorphic dataset. 
By leveraging on the ACES platform, the taxels of the sensing array are capable of emulating the function of the fast-adapting (FA) mechano-receptors of the human skin.  Similar to their biological counterparts, the taxels respond only to dynamic pressure (i.e., dynamic skin deformation) but are insensitive to static forces~\cite{johansson2009coding}. 

With the ACES platform, each taxel captures and transmits FA responses  asynchronously as ``events'' using spikes (i.e., electrical pulses), and that data is sent by individual taxels only when necessary. This is achieved by encoding the taxels with unique electrical pulse signatures. These signatures are designed to overlap robustly for deconvolution, such that multiple taxels are allowed to transmit information without specific time synchronization. Moreover, this encoding scheme permits all signatures from the stimulated taxels to be combined and propagated to the decoder via one electrical conductor, thus yielding a lower readout latency and simpler wiring. A decoder then deconvolves and identifies the stimulated taxels to retrieve the spatio-temporal tactile information by correlating the received (combined) signatures against the encoded ones. In the following subsection, we describe the data collection process using this neuromorphic tactile system.

\subsection{Data Collection} \label{datacollection}
\begin{figure}
	\centering
	\includegraphics[scale=0.395]{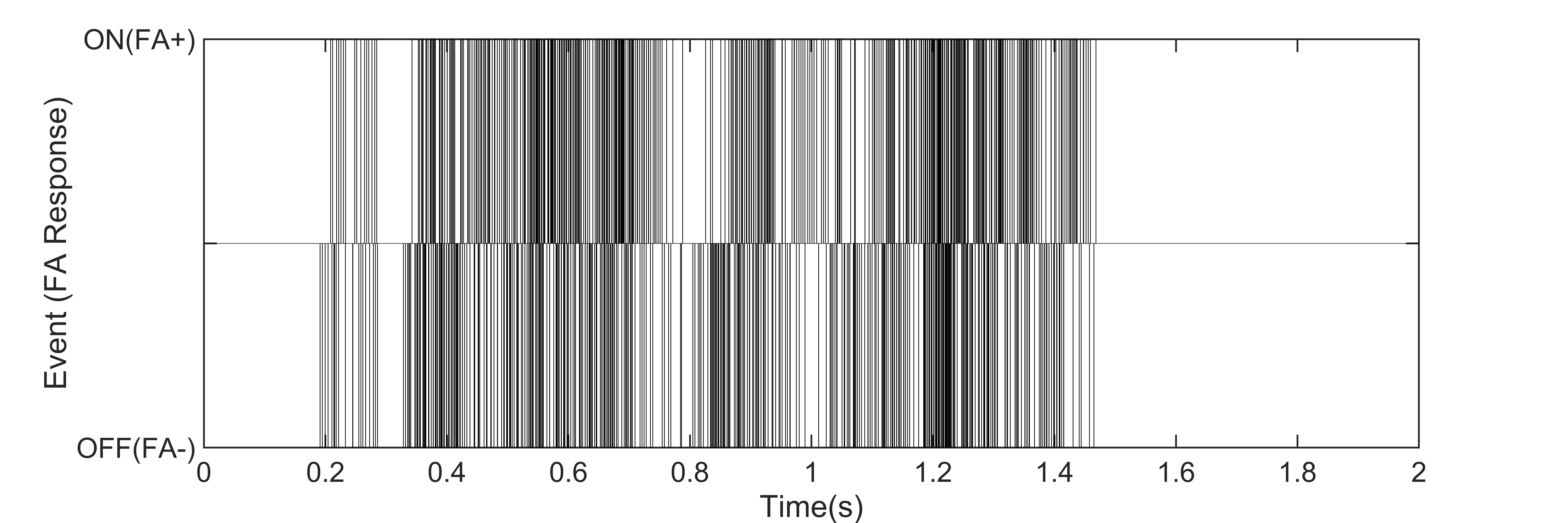}
	\caption{Decoded FA responses (i.e., events) of the tactile sensor array for Digit 8.}
	\label{fig:combinedevent}
\end{figure}

Twenty-three participants were asked to use a pen (tip diameter: 0.5 mm) to write within the tactile-sensing array one digit at a time. This process was repeated 30 times to collect 30 samples of each digit from an individual writer. The combined signal output (i.e., the handwriting data) from the 100 taxels of the tactile sensor array was sampled at 63 MHz utilizing an oscilloscope (Picoscope 3406D) for 2 seconds. The FA responses (i.e., events) were decoded offline in MATLAB, and one of the representative results is depicted in Fig. \ref{fig:combinedevent}.
We collected a total of 6,953 samples from 23 participants. The following subsection provide the details on how to access and use the resulting dataset.

\subsection{File Formats} \label{fileformat}
\begin{figure}
	\centering
	\includegraphics[scale=0.0475]{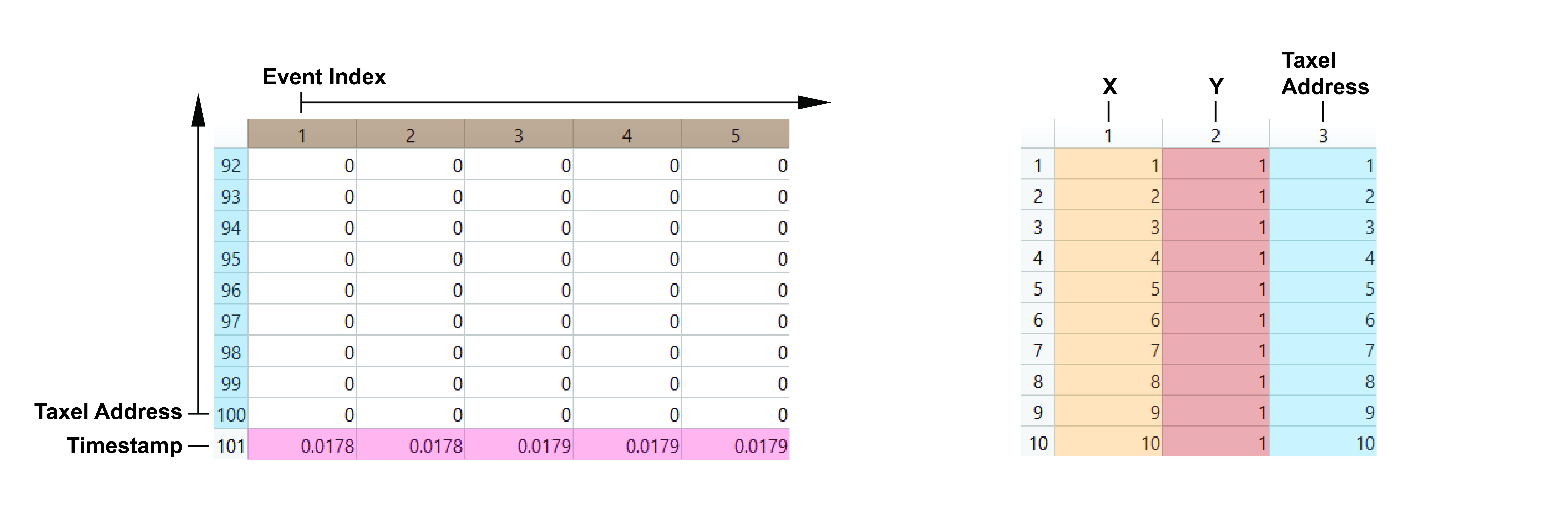}
	\caption{Snippets of (\textbf{Left}) the \textit{spiketrain} array and (\textbf{Right}) the lookup table.}
\label{fig:SpikeArr}
\end{figure}

The full ST-MNIST dataset can be accessed online\footnote{ \url{http://www.benjamintee.com/stmnist}}. A separate directory exists for each digit, and each sample is saved as a .MAT file which contains an array named \textit{spiketrain} (Fig. \ref{fig:SpikeArr}, left). This array has a size of 101 $\times$ $n$, where $n$ is the total number of events in the sample. The FA responses captured by the individual taxels are appended to the rows of \textit{spiketrain} according to the taxel addresses (which are represented by the row indexes). The increase(FA$+$) and decrease(FA$-$) of the pressure exerted on the taxels are denoted by 1(ON) and  $-$1(OFF) in the \textit{spiketrain}, respectively, with 0 for no event. The last row of {\textit{spiketrain}} provides the event timestamps (in seconds). A lookup table (i.e., LUT.mat) is included to map the taxel addresses to the X and Y addresses of the tactile sensor array (Fig. \ref{fig:SpikeArr}, right). 

\section{Results} \label{results}
\subsection{Dataset Properties} \label{data_properties}

Table {\ref{table:statistics}} presents some basic properties of the ST-MNIST dataset. The large standard deviations (SDs) of the number of ON and OFF events for the individual classes are attributed to the different handwriting styles of the participants. All classes are also observed to have large SDs for the range of X and Y addresses because no spatial restriction was imposed on the pen-down and pen-up points. Notably, this is not the case for N-MNIST~\cite{orchard2015converting},  which was converted from the static images of MNIST by using a neuromorphic camera. The range of X and Y addresses for N-MNIST is only dependent on the image size (which is the same for all MNIST images), and hence, their SDs are 0.

The statistics in Table \ref{table:statistics} are shown to be highly dependent on the handwritten digits, and can therefore be used for classification. However,  prior work~\cite{orchard2015converting} suggests that utilizing these statistical data as features for a k-Nearest Neighbor classifier are sub-optimal for classification. Therefore, we  focus on employing the spike-trains (instead of just the statistics) for classification.

\begin{table}[h!]
	\centering
	\caption{Statistics of ST-MNIST.}
	\scalebox{0.8}{
	\begin{tabular}{C{1.85cm}|C{2.8cm}|C{2.6cm}|C{2.6cm}|C{2.7cm}|C{2.5cm}}
		\hline
		\hline
		&\multicolumn{5}{c}{\textbf{Digit}}\\
		\cline{2-6}
		&\textbf{0}&\textbf{1}&\textbf{2}&\textbf{3}&\textbf{4}\\
		\hline
		\textbf{Statistic}&\multicolumn{5}{c}{\textbf{Mean(SD)}}\\
		\hline
		ON events 	&  1738.13 (1365.54)  & 362.95 (204.57) &  878.78 (405.13)  & 1053.44 (673.43) & 823.63 (458.51) \\ 
		\hline
		OFF events 	&  1139.66 (803.17)  & 245.77 (124.34) &  623.88 (294.57)  & 810.22 (1403.10) &  597.57 (355.25)  \\
		\hline
		X mean 		&  5.57 (0.51)  & 6.17 (0.70) &  6.57 (0.70)  & 5.88 (0.59) &  6.05 (0.57)  \\
		\hline
		Y mean 		&  6.84 (0.75)  & 6.21 (1.16) &  7.02 (0.82)  & 7.12 (0.88) &  6.93 (0.93) \\
		\hline
		X range 	&  8.92 (0.53)  & 8.72 (0.70) &  8.80 (0.64)  & 8.94 (0.53) &  8.95 (0.52)  \\
		\hline
		Y range 	&  8.48 (1.04)  & 7.04 (1.36) &  8.21 (1.04)  & 8.20 (1.08) &  8.44 (0.87)  \\
		\hline
		Sample size &  697  & 690 &  699  & 690 &  690 \\
		\hline
		\hline
		&\multicolumn{5}{c}{\textbf{Digit}}\\
		\cline{2-6}
		&\textbf{5}&\textbf{6}&\textbf{7}&\textbf{8}&\textbf{9}\\
		\hline
		\textbf{Statistic}&\multicolumn{5}{c}{\textbf{Mean(SD)}}\\
		\hline
		ON events 	&  1113.90 (701.60)  & 739.37 (521.99) &  809.30 (631.05)  &  1399.14 (898.56) &  773.13 (483.25) \\ 
		\hline
		OFF events 	&  819.76 (677.03)  & 548.48 (410.00) &  538.98 (361.39)  & 1000.42 (755.04) &   546.29 (295.75)  \\
		\hline
		X mean 		&  5.41 (0.53)  & 6.04 (0.45) &  5.32 (0.57)  & 5.78 (0.57) &   5.24 (0.54) \\
		\hline
		Y mean 		&  6.57 (0.79)  & 6.68 (0.76) &  6.92 (0.89)  & 6.76 (0.90) &  6.58 (0.91) \\
		\hline
		X range 	&  8.91 (0.58)  & 8.94 (0.52) &  8.96 (0.49)  & 8.94 (0.52) &  8.91 (0.53)  \\
		\hline
		Y range 	&  8.38 (1.00)  & 8.33 (1.02) &  7.97 (1.18)	  & 8.19 (1.17) &  8.17 (1.13)  \\
		\hline
		Sample size &  693  & 698 &  704  & 697 &  695 \\
		\hline
		\hline
	\end{tabular}
		}
	\label{table:statistics}
\end{table}

\begin{figure}
	\centering
	\includegraphics[scale=0.55]{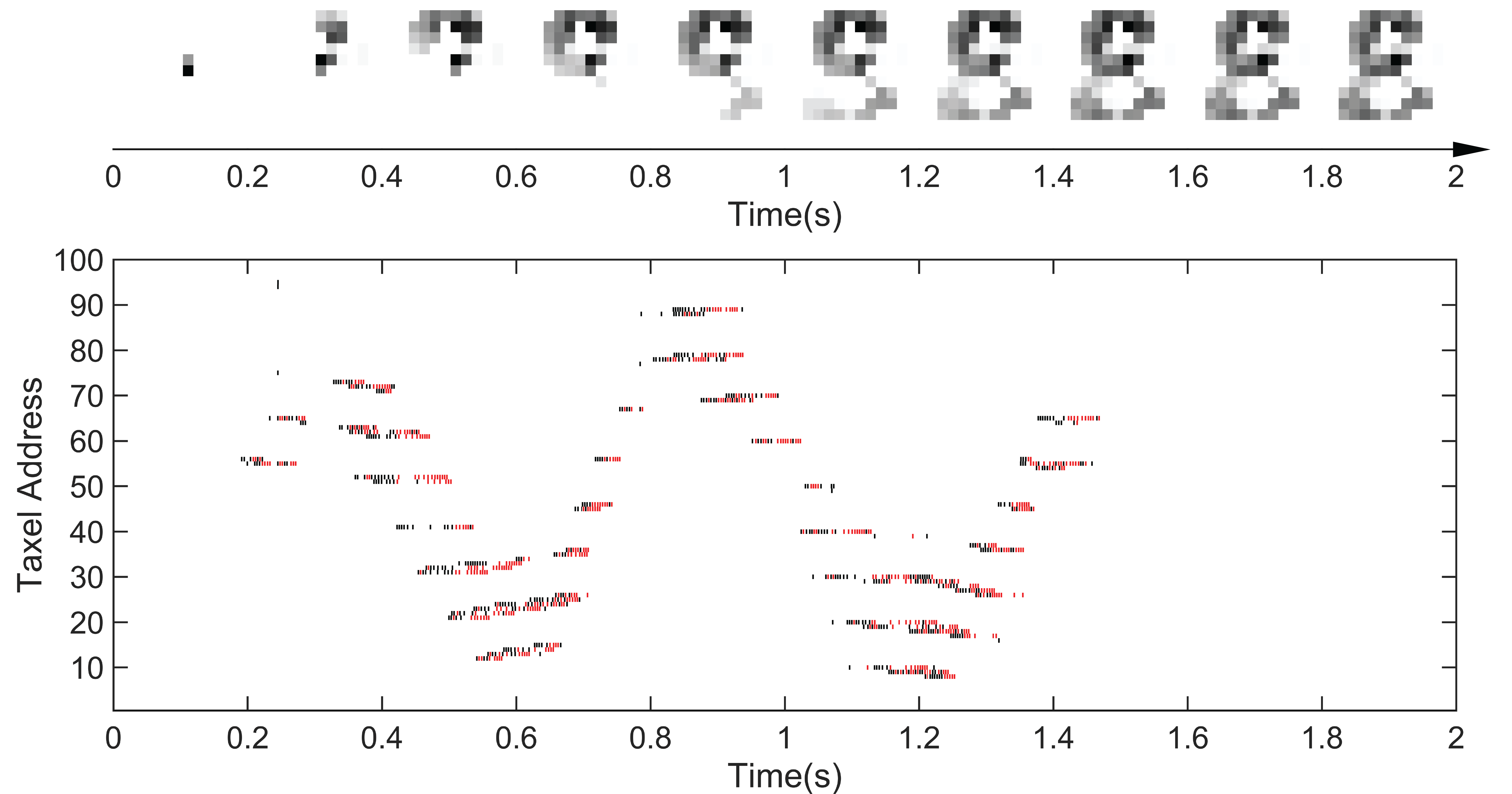}
	\caption{(\textbf{Top}) Evolution of the normalized ``tactile image'' over time for Digit 8. (\textbf{Bottom}) Spike raster plot of the FA responses (i.e., events) of the 100 taxels for 2 seconds. The ON events are in red, and OFF events are in black.}
	\label{fig:recordings}
\end{figure}

\subsection{Classification}  \label{classification}

Here we briefly present the classification accuracies of the ST-MNIST dataset to serve as performance benchmarks for future research. We acquired these accuracies using the existing ANN~\cite{krizhevsky2012imagenet} and SNN~\cite{hunsberger2016training,shrestha2018slayer} models; the models were applied without modification. We refer the reader to the original papers for the detailed description of the models, and survey articles for the background on SNNs~\cite{Pfeiffer2018} and ANNs~\cite{lecun2015deep}.

Three approaches were implemented to classify the handwritten digits and each method was repeated 10 times on random train-test (80-20) splits. The first employs a Convolutional Neural Network (CNN) model~\cite{krizhevsky2012imagenet}, which is a conventional deep ANN. The second uses a SNN model that is converted from a trained ANN (ANN-SNN)~\cite{hunsberger2016training}. The last utilizes a SNN model trained with SLAYER~\cite{shrestha2018slayer}, a method of error backpropagation for SNNs.
The implementation of these approaches is described in the subsections below.

Throughout this paper, the following notation is used to indicate the model architecture; this notation is adapted from \textit{Shrestha et al.}~\cite{shrestha2018slayer}. The  model layers are separated by $-$, and the spatial dimensions are separated by $\times$. A convolutional layer is denoted by c, and an output neuron is indicated by o. For instance, a model architecture of 10$\times$10$-$9c3$-$8o indicates that it has an input of 10$\times$10, followed by 9 convolutional filters of 3$\times$3, and lastly a dense layer of 8 output neurons.

\subsubsection{Convolutional Neural Network (CNN)} 

CNNs have been applied with great success to face recognition~\cite{taigman2014deepface}, and the detection of objects in images~\cite{cirecsan2012multi}. They are designed to process data that come in the form of arrays (such as, a grayscale image composed of a single 2D array containing pixel intensities). Hence, the spike data of ST-MNIST need to be pre-processed to form arrays for the implementation of CNN.
We pre-processed the dataset to form ``tactile images'' (Fig. \ref{fig:combinedevent}, top panel) by aggregating the FA responses. Each ``tactile image'' is a 10$\times$10 array, where each element corresponds to the aggregated FA responses from a single taxel. The "tactile image" is then normalized and used as an input to the CNN for classification. The CNN architecture is shown in Table \ref{table:class_results}, which includes 2 convolutional layers (with Rectified Linear Unit as the activation function) and an output layer (with Softmax as the activation function). We developed the model in Python using TensorFlow, and it was trained for 20 epochs to minimize the sparse categorical cross-entropy loss using the RMSProp optimizer. 

\subsubsection{Conversion of ANN to SNN (ANN-SNN)}\label{subsec:conversion}

SNNs are inspired by biological neural networks, where asynchronous spike signals are processed in a massively parallel fashion~\cite{Pfeiffer2018}. Hence, they are well suited for processing spatio-temporal event-based information from neuromorphic sensors. The asynchronous event-driven mode of computing in SNNs has led to lower power consumption, and faster inference. To best exploit these advantages in practice, SNNs should be run on neuromorphic hardware (e.g., Intel Loihi~\cite{davies2018loihi} and IBM TrueNorth~\cite{merolla2014million}). However, SNNs are still limited by the lack of good training algorithms.
Conventional deep learning has relied on stochastic gradient descent and error backpropagation, which are not applicable for the non-differentiable spiking neurons. Therefore, the integration of these spiking neurons into the gradient-based training process would require additional effort. 
To circumvent the issues of gradient descent in SNNs, a conventionally trained ANN can be converted into SNN~\cite{hunsberger2016training}.
Here, we translated a CNN into SNN by training the model with the rate-based implementation of leaky integrate-and-fire (LIF) neurons, and then replacing these neurons with the actual spiking ones for inference. The model architecture is shown in Table \ref{table:class_results}, which comprises an input layer (with ``tacile image'' as the input), 2 convolutional layers and an output layer. We developed the model in Python using TensorFlow and NengoDL, and it was trained for 20 epochs to minimize the sparse categorical cross-entropy loss using the RMSProp optimizer.
Furthermore, we also implemented the converted SNN (with a set of 200 random test samples) on the Intel Loihi neuromorphic chip~\cite{davies2018loihi} via the Intel Neuromorphic Research Community (INRC) cloud service for inference. 

\subsubsection{SNN with SLAYER} 

Here, we implemented a SNN trained with SLAYER~\cite{shrestha2018slayer}. As discussed in Sec. \ref{subsec:conversion}, the non-differentiable spiking neurons have prohibited the direct application of backpropagation techniques to SNNs. SLAYER resolves this issue by using (i) a stochastic spiking neuron approximation to approximate the derivative of the neuron model, and (ii) a temporal credit assignment policy to backpropagate errors. The SNN architecture is shown in Table \ref{table:class_results}, and it has an input size of 10$\times$10$\times$2 (with a positive and negative polarity channel per taxel). We implemented the SNN in Python using PyTorch, and the Spike Response Model (SRM)~\cite{gerstner1995time} is used to model the spiking neurons. The SNN was trained for 100 epochs with SLAYER to minimize the differences between the observed and desired spike counts for the output neurons~\cite{shrestha2018slayer}. The output neuron that generates the highest spike count represents the winning class. For the training process, the spike data of ST-MNIST are binned into fixed-width intervals of 0.01s (200 bins), and the desired true and false spike counts were set to 60 and 10 respectively.

\begin{table}[h!]
	\centering
	\caption{Classification Results of ST-MNIST.}
	\scalebox{0.8}{
		\begin{tabular}{C{5cm}|C{6cm}|C{3.5cm}}
			\hline
			\hline
			\textbf{Model}&\textbf{Architecture}&\textbf{Mean Accuracy(SD)}\\
			\hline
			CNN~\cite{krizhevsky2012imagenet}   & 10$\times$10$-$6c4$-$24c3$-$10o & 0.89 (0.01)\\
			\hline
			ANN-SNN~\cite{hunsberger2016training}  & 10$\times$10$-$6c4$-$24c3$-$10o \newline SNN converted from standard ANN  & 0.87 (0.01)\\
			\hline
			
			ANN-SNN~\cite{hunsberger2016training}  \newline(Intel Loihi-INRC)  & 10$\times$10$-$6c4$-$24c3$-$10o \newline SNN converted from standard ANN  & 
			
			0.86 (0.02)\\
			\hline
			SNN-SLAYER~\cite{shrestha2018slayer}& 10$\times$10$\times$2$-$6c2$-$24c3$-$10o &0.81 (0.01)\\ 
			\hline
			\hline
		\end{tabular}
	}
	\label{table:class_results}
\end{table}

\subsubsection{Classification Results} 

Classification accuracies of three methods on the ST-MNIST dataset are presented in Table \ref{table:class_results} with standard deviations (SDs) shown in brackets. The results show that the ANN-SNN model has suffered a loss of 2\% in accuracy in comparison to CNN. Despite the drop in accuracy, the ANN-to-SNN conversion approach still has its advantages. The main highlight of this approach is that the full toolkit of deep learning techniques can be exploited, meaning that state-of-the-art deep ANNs for classification tasks can be translated into SNNs~\cite{Pfeiffer2018}. Furthermore, the converted SNN can also be implemented on the highly energy-efficient neuromorphic hardware, and it yields good classification accuracy as shown in Table \ref{table:class_results}. However, both CNN and ANN-SNN models require pre-processing on the spike data to form ``tactile images'' as their inputs, and this may incur additional latency. In contrast, the spike data can be directly fed into the SNN-SLAYER model for classification, and its accuracy is 81\%.

\section{Conclusion} \label{conclusion}

In this work, we present ST-MNIST, a novel neuromorphic tactile dataset comprising  human-generated 10-class handwritten digits. It was generated from a recent tactile sensing hardware (i.e., ACES~\cite{lee2019neuro}) that uses an asynchronous network of biomimetic touch receptors. There are efforts by others (e.g., MNIST-DVS\footnote{http://www2.imse-cnm.csic.es/caviar/MNISTDVS.html} and N-MNIST~\cite{orchard2015converting}) to convert MNIST images to neuromorphic data by using event cameras. To our best of knowledge, the ST-MNIST dataset described here represents the first and largest publicly available neuromorphic tactile dataset to date. We hope that this novel dataset will spur even greater research interest from the growing neuromorphic community and accelerate machine learning algorithm development. 

In this paper, we also included an initial effort to provide a few classification accuracies of the ST-MNIST dataset. These accuracies were obtained using the existing ANN~\cite{krizhevsky2012imagenet} and SNN~\cite{hunsberger2016training,shrestha2018slayer} models, and aim to serve as performance benchmarks for future research. We did not modify or optimize the models, therefore the results presented in Table \ref{table:class_results} could be regarded as the minimum classification accuracies upon which to improve. Nevertheless, there still exists room for improvement and this is especially true for SNNs. Historically, SNNs have been limited by the lack of good training algorithms that make specific use of the capabilities of spiking neurons. Consequently, their classification accuracies on typical benchmarks (such as MNIST~\cite{lecun1998gradient}) do not consistently outperform ANNs. To some extent, the underperformance of SNNs could be attributed to the nature of those benchmarking datasets, which are based on conventional frame-based datasets.
We believe the benefits of SNNs for machine learning tasks on neuromorphic sensor data have yet to be fully explored. We hope that our ST-MNIST dataset will encourage research in this area to exploit spatio-temporal event-based data for SNNs in a way that makes them competitive with deep ANNs.

\bibliographystyle{unsrt}  
\newpage
\bibliography{references} 
\end{document}